\documentclass[conference]{IEEEtran}
\IEEEoverridecommandlockouts
\usepackage{cite}
\usepackage{amsmath,amssymb,amsfonts}
\usepackage{algorithmic}
\usepackage{graphicx}
\usepackage{textcomp}
\usepackage{xcolor}
\usepackage{booktabs}
\usepackage{cleveref}
\def\BibTeX{{\rm B\kern-.05em{\sc i\kern-.025em b}\kern-.08em
    T\kern-.1667em\lower.7ex\hbox{E}\kern-.125emX}}

\usepackage[numbers]{natbib}
\usepackage{enumitem,amssymb}
\newlist{todolist}{itemize}{2}
\setlist[todolist]{label=$\square$}
\usepackage{pifont}

\usepackage{todonotes}
\usepackage{caption}
\usepackage{subcaption}

\begin{document}

\title{\huge SMaRCSim: Maritime Robotics Simulation Modules  
\thanks{}
}

\author{Mart Kartašev, David Dörner, Özer Özkahraman, Petter Ögren, Ivan Stenius, John Folkesson
\thanks{The authors are with the KTH Royal Institute of Technology, Stockholm, Sweden}
}
\maketitle

\begin{abstract}
Developing new functionality for underwater robots and testing them in the real world is time-consuming and resource-intensive.
Simulation environments allow for rapid testing before field deployment.
However, existing tools lack certain functionality for use cases in our project: i) developing learning-based methods for underwater vehicles; ii) creating teams of autonomous underwater, surface, and aerial vehicles; iii) integrating the simulation with mission planning for field experiments.
A holistic solution to these problems presents great potential for bringing novel functionality into the underwater domain.
In this paper we present SMaRCSim, a set of simulation packages that we have developed to help us address these issues.
\end{abstract}


\section*{Software Release:}
The SMaRC simulation modules can be found at \texttt{https://github.com/smarc-project/smarc2}

\section{Introduction}

Maritime Robotics is a multidisciplinary field requiring comprehensive solutions to a variety of technical challenges; Sensing, dynamics and control, mapping and localization.
In addition to each being an entire research field by itself, field experiments also require special attention due to the inherent challenges and risks involved. 
When it comes to simulation, all these aspects require various simulation features that do not always intersect.  

\begin{figure}
    \centering
    \begin{subfigure}[b]{\linewidth}
         \centering
         \includegraphics[width=\textwidth]{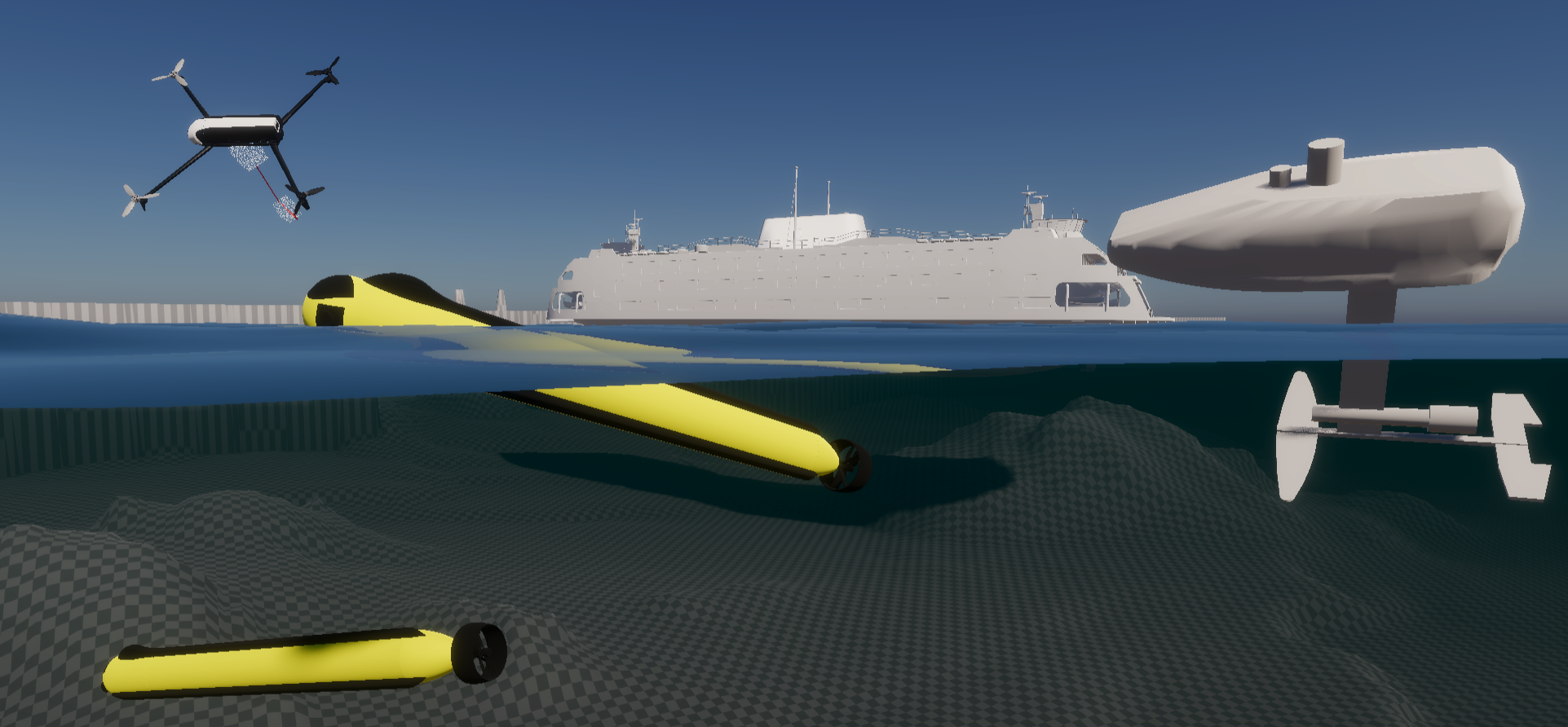}
         \caption{A quadrotor drone, the SAM AUV, a hydrofoiling vehicle, and a surface vessel, all simulated within the same scene with SMaRCSim.}
         \label{fig:multi}
     \end{subfigure}
     \hfill
         \begin{subfigure}[b]{\linewidth}
         \centering
         \includegraphics[width=\textwidth]{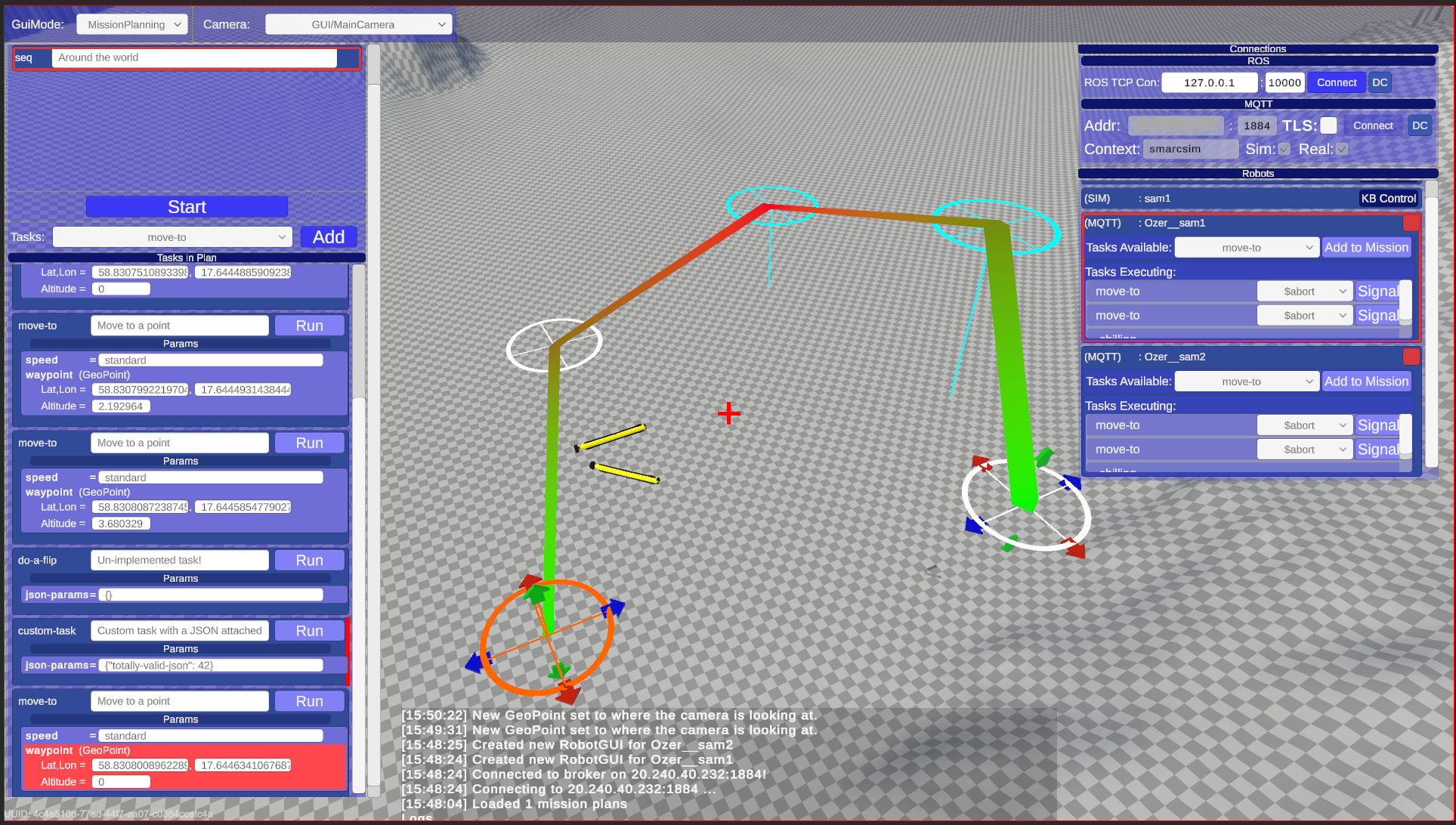}
         \caption{SMaRCSim C2 interface. Water rendering can be turned off, to show all waypoints and agents in the world more clearly.}
         \label{fig:c2}
     \end{subfigure}
     \hfill    
    \begin{subfigure}[b]{\linewidth}
        \centering
        \includegraphics[width=\textwidth]{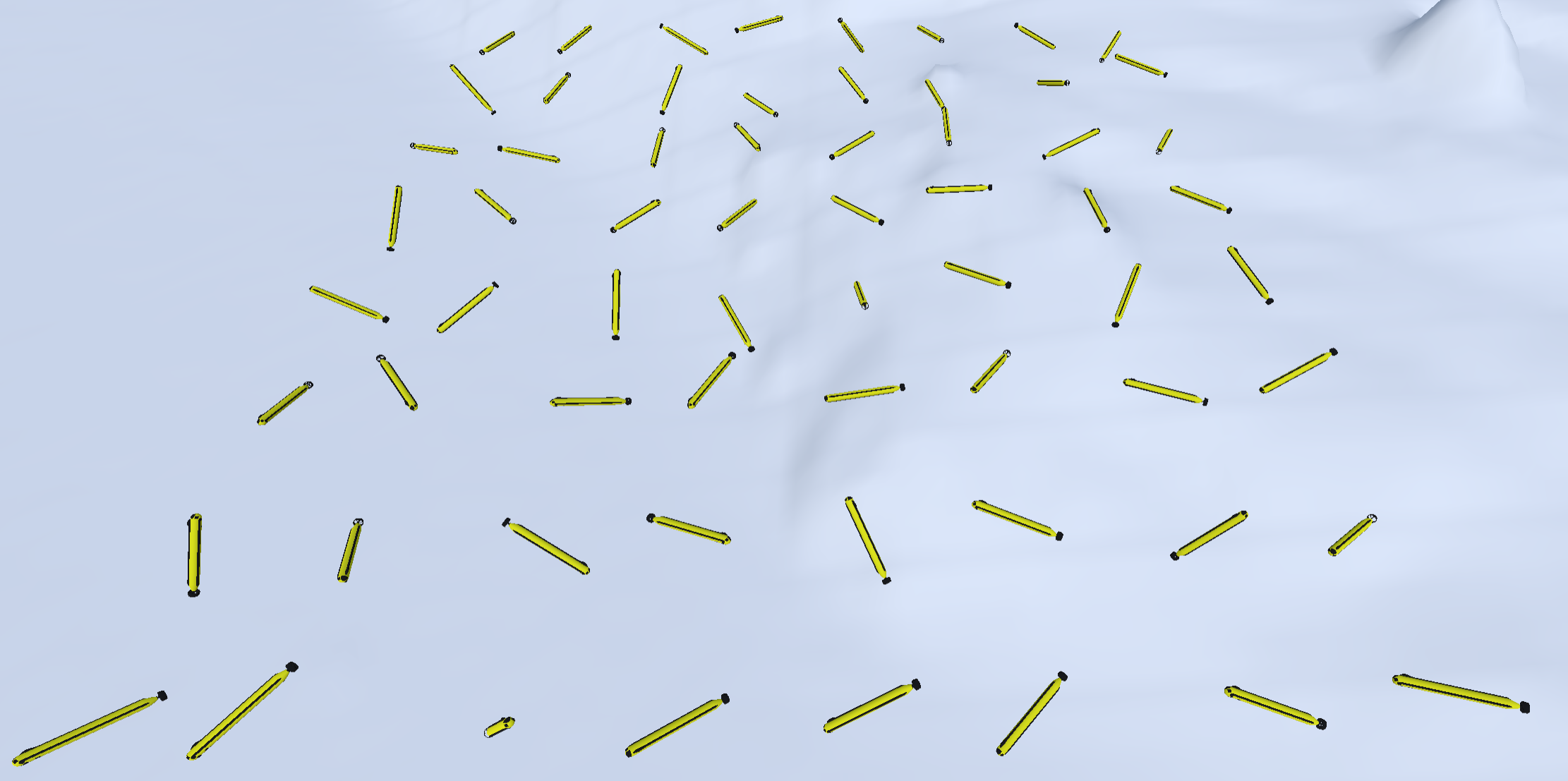}
        \caption{64 AUVs running faster than real-time to generate experiences for RL.}
        \label{fig:rl}
    \end{subfigure}
    \caption{Examples of the three application scenarios.}
    \label{fig:pretty_pictures}
    \vspace{-0.6cm}
\end{figure}

While there are already numerous different maritime underwater simulators, each tends to focus on a particular aspect of maritime robotics. 
Although their individual focus can be beneficial when it overlaps with the users', it can be limiting for projects that aim to develop and verify a system in its entirety.
In SMaRC\footnote{\texttt{https://smarc.se/}}, we operate across several disciplines and thus need a simulation platform that is easily extensible for each new research area.
We also require the simulation package to be easy to learn and use so that new researchers can focus on their own topic instead of the simulator itself.

This paper aims to give an overview of the ongoing work on simulation at SMaRC. 
More precisely, we describe the various tools we have developed using the following scenarios:
\begin{itemize}
    \item Multi-domain, Mixed-fidelity projects, which focus on missions with a mix of vehicles in different domains and levels of simulation.
    \item Mission planning, which discusses the need to support field operations through planning, data processing and simulation.
    \item Controller development and machine learning, emphasizing customizable physics models, Sim-2-Real work, faster than real-time simulation, and integration with scientific libraries.
\end{itemize}

\begin{table*}[]
    \centering
    \caption{Comparison of the latest underwater simulator capabilities. $^*$Maritime includes underwater and surface vessels. $^{\dagger}$based on the respective GitHub repo, if available at time of writing (Feb. 18, 2025). $^{\ddag}$Repositories not accessible without an account.}
    \label{tab:sim_comparison}
    \begin{tabular}{ lccccccc }
     \toprule
     Feature & SMaRCSim  & MARUS & HoloOcean & Stonefish & UUVSim & DAVE & UNav Sim\\
     \midrule
     Domain &  Maritime$^*$, Aerial  & Maritime$^*$ & Underwater& Underwater & Underwater & Underwater & Underwater\\
     ROS Support &  ROS 2 & ROS 1, 2 & ROS 1 & ROS 1 & ROS 1 & ROS 1 & ROS 1, 2\\
     Framework & Unity & Unity & Unreal & Custom & Gazebo & Gazebo & Unreal\\
     Physics Engine & PhysX & PhysX & PhysX, Chaos & Bullet & Bullet-based & Bullet-based & Chaos\\
     Latest update$^{\dagger}$ & 2025 & 2023 & Inaccesible$^{\ddag}$ & 2024 & 2020 & 2023 & 2024\\
     Integrated C2 & Yes & No & Non-graphical & No & No & No & No\\ 
     RL feasible & Yes ($\sim$50xRT) & Partial & Partial ($\sim$2xRT) & No & No  & No & No \\

    \bottomrule
    \end{tabular}
\end{table*}

For reasons that will become apparent, we chose Unity as our simulation tool.
It allows us to build, prototype, and modify systems in a unified environment. 
Additionally, it offers integration with work from other providers, enabling us to reuse existing resources effectively. 
Unity's ease of onboarding also ensures that our constantly changing academic team can quickly learn and contribute to the project.
In addition, it supports multiple communication protocols such as ROS (1 and 2) and MQTT, which are indispensable for holistic testing at different levels of system maturity.

\section{Related Works}

Simulators are crucial for developing control methods and conducting mission planning for underwater robotics. 
Bhat et al. presented such a full mission stack in \cite{bhat.2020a} where Neptus \cite{pereira.2006} is used for mission planning and Stonefish  \cite{cieslak.2019} for simulation.
Stonefish provides a ROS \cite{macenski.2022} interface, making it suitable for hardware-in-the-loop (HIL) testing.
However, it has not been updated to support ROS 2.
Stonefish provides some hydrodynamic modeling based on the Bullet Physics engine (taking inspiration from UWSim \cite{prats.2012}), which is also open to modification by users due to its open-source nature.
However, we have found such modifications to require significant effort to achieve.
In addition to its dynamics, Stonefish provides high visual fidelity while aiming for real-time simulation of one vehicle.
This makes it unsuitable for learning and multi-agent approaches.

Some alternatives to Stonefish include the Gazebo-based UUVSim \cite{manhaes.2016} (which we have also used before Stonefish \cite{bhat.2019}), DAVE \cite{zhang.2022} and WAVE \cite{rosette.2024}.
While they are each very suitable for the focus area for which they were created, they are all unmaintained and lack support for ROS 2.

Game engines such as Unity \cite{juliani2020} or Unreal \cite{unrealengine} have also been used.
These engines can produce executables for many operating systems and platforms, freeing the users from using a specific operating system.
URSim \cite{katara.2019}, and UWRoboticsSim \cite{chaudhary.2021} are built with Unity but have not been maintained.
The latest Unity-based simulator is MARUS \cite{loncar.2022}, which features an extensive list of sensors, as well as underwater and surface vessels, with a focus on vision-based algorithms. 
MARUS also has ROS 2 support. 
However, some of MARUS is proprietary which hinders open and reproducible research.

Another game engine that could be used is the Unreal Engine, used by HoloOcean \cite{potokar.2022} and UNav Sim \cite{amer.2023}.
HoloOcean provides high-fidelity visuals as well as an interface to a set of reinforcement learning algorithms. 
However, they only have single-agent ROS 2 support, as well as proprietary scenes, making modifications challenging. 
Like HoloOcean, UNav Sim focuses on high-fidelity visuals. 
Furthermore, they provide a ROS 2 interface to validate planning and control algorithms. 
A common point for simulators based on the Unreal Engine is that visual fidelity is given a very high priority.

Beyond that, there are vehicle simulators built in MATLAB/Simulink such as Simu2Vita \cite{decerqueiragava.2022} as well as the MATLAB \cite{fossen.2025a} and Python simulator \cite{fossen.2025} provided by Fossen et al. to complement \cite{fossen.2021}. 
However, these simulations only focus on the dynamics of specific vehicle types and are therefore less useful for larger scenarios.
An overview of a set of recent simulators can be found in \cref{tab:sim_comparison}.

\section{Application scenarios}

There is a wide variety of tools available, with specific trade-offs being made in each one.
In this section, we highlight some scenarios where we have found existing solutions lacking for our purposes.
    
\subsection{Multi-Domain, Mixed-Fidelity Projects}
Most existing simulators target specific types of vehicles or domains.
When examining scenarios with vehicles that are in multiple domains, these single-domain simulators are insufficient. 
For example, simulation of an Unmanned Aerial Vehicle (UAV) recovering and delivering an Autonomous Underwater Vehicle (AUV) to the shore is not possible in a simulator that assumes everything is underwater. 
Such a simulator requires sufficiently realistic underwater, surface, air and land environments and vehicles to be simulated simultaneously.

A common use case in research is to simulate a vehicle only partially, or to simulate its parts at different levels of fidelity.
The ability to have different levels of abstraction in simulation allows research to be done in parts or even in parallel.
For example, developing an obstacle avoidance algorithm for a hydrofoiling vehicle where the vehicle dynamics might not need to be simulated to the same level as the sensors.

Although the common point between these scenarios is a requirement of flexibility, that is not sufficient. 
Open-source simulators allow one to completely modify them freely at the cost of significant time and knowledge.
In our experience, modifying simulators through source code has usually been infeasible within the constraints of academic work.
Thus, the simulator internals must also be accessible with little time investment.
By using Unity, we leverage its visual editor and the plethora of available tutorials and examples, which keeps it accessible to new researchers.
The prefab system provides rapid modification and construction of new vehicles and environments, using existing objects as building blocks that work out of the box.
Through its interface to its physics engine, Unity also allows deeper modification of the simulation.

\subsection{Mission Planning}
Although there are simulators that achieve high fidelity in very narrow scenarios, research on novel vehicles, sensors, and control systems still requires holistic tests that can only be conducted in reality.
Field tests require a  \emph{command and control} (C2) software that can adapt to the changing research topics just as well as the simulator.

Mission planning is usually the first step of field testing.
Each unique vehicle has a unique set of capabilities that can be planned for, which the C2 software must make available.
In our experience \cite{bhat.2019, bhat.2020a}, existing C2 software lacks the flexibility to easily create new maneuvers and actions for a new type of vehicle.
In our solution, we use the WARA-PS API \footnote{\texttt{https://api-docs.waraps.org/}} designed for multi-domain C2.
We have developed a user interface that allows interaction with this extensible, multi-domain API.

Most C2 software presents a flat, top-down view, working on top of a 2D map.
While this is sufficient for most surface and land based missions, the depth and height dimension is important for underwater and air missions, thus a 3D-capable mission planner is needed.
This capability is also useful for surface and land missions, since it allows better visualization of collected spatial data (LiDAR, RGB-D).
Unity is primarily a 3D-first game engine and thus fulfills this need out of the box.
Our mission planning interface makes use of in-world representations of tasks and parameters where applicable, such as waypoints and paths.

An inconvenient aspect of the underwater domain is that satellite maps are usually not sufficient due to the seabed topography being obscured from space beyond a certain depth.
This makes it difficult to plan missions for vehicles that need to get close to the sea bottom.
Oftentimes, before such a mission, a survey is conducted to collect bathymetry data.
This newly collected data can be critical to the success of a mission, therefore the C2 software must be able to display it in the field.
Since such experiments are often done in remote locations, C2 software must do so without relying on a remote connection.
Through many available libraries, virtually all formats of data can be imported into Unity.
Many common formats (such as OBJ, DXF, DAE) are natively supported, and can be made part of a scene by simply dragging the file into the Unity window.
Thus, after a survey, the produced bathymetry data can be imported into Unity with little effort and used for C2 right away.

The second part of C2 is commanding the vehicle, for which the state of the vehicle needs to be known.
During underwater operations, continuous connection to the vehicle is usually not available due to limited bandwidth and high energy use of acoustic modems.
This means the C2 software can only receive small, infrequent updates of vehicle state.
In between updates, researchers should be able to simulate what the vehicle \emph{might} be doing, so that they can make decisions and issue commands as needed.
Merging of C2 and simulation also leads to decreased training needs, leading to quicker adoption and increased ease of use in the field.
Since Unity has libraries to support external connections (specifically ROS 2 and MQTT), updates from the real vehicle can be applied to simulated vehicles in our solution.

\subsection{Controller Development and Sim-2-Real-2-Sim}

Controller and dynamical model development with analytical models like Model-Predictive Control and learning-based approaches like Reinforcement Learning (RL) highly benefit from faster-than-real-time simulations.
In addition to speed, simultaneous access to the simulation loop and scientific libraries is invaluable.
Unity provides the MLAgents \cite{juliani2020} platform, which includes a Low-Level API, which in turn provides a direct interface to Python where most RL and scientific libraries are available.
Unity supports simulation speeds of up to 100x real-time\footnote{Subject to CPU resources and complexity of the particular simulation.} natively, as a simple configuration.

A significant concern with RL is bridging the Sim-2-Real gap for the simulation-trained policies to be feasible in the real world. 
Although game engines are powerful and highly optimized, they often lack certain forces, such as those created by control surfaces. 
To add these missing effects, the simulation engine must allow the application of manually computed forces in addition to the base simulation.
Unity provides access to its underlying physics engine and rigid body objects, which have allowed us to implement such forces for our vehicles.
We have used the same mechanisms to also produce external effects on the vehicles, such as winds and currents.

Over time, improving the sim's accuracy proved useful for research and development of more than just RL, which has led to a more general drive for improvement of dynamical fidelity. 
We have described how it is possible to improve the dynamics of the engine. 
Without knowing how to parameterize the various complementary forces accurately, our results remain heuristic at best. 
Various methods, such as classical System Identification (SI), Physically Informed Neural Networks (PINNs), Gaussian Processes (GPs), and more, are being evaluated to model so-called residuals that, in simple terms, try to measure and correct the error between the real world and the simulation. 
This alternative to building a full vehicle model is inspired by recent works on drones \cite{kaufmann_champion-level_2023}. 
Such data-oriented solutions require access to accurate ground-truth data. 
Currently, a pool for underwater experimentation, equipped with a motion capture system, is being built at KTH, which will help with such approaches. 
For such data to be helpful from a simulation perspective, software solutions for projecting real-world vehicles back into the simulation (Real-2-Sim) - based on both ROS2 and the direct LLAPI - are also being developed. 
Having data flow in both directions between the real world and the simulation defines the Sim-2-Real-2-Sim capability, which is currently one of the primary goals for development.

\section{Conclusion}
In this paper, we presented the ongoing work on SMaRCSim, our collection of modules for maritime robotics simulation based on the Unity game engine.
SMaRCSim covers multiple domains, from under the water into the air, at different fidelities as desired.
Through systems available in Unity, we can rapidly develop new environments and vehicles and keep them updated.
Additionally, the ease of use, combined with the massive existing user base, makes it very accessible for students and incoming researchers to dive in and develop their ideas quickly. 
Our integrated simulation and command and control interface allows the development and testing of novel algorithms both before and during deployment in unstructured, real-world environments. 
SMaRCSim enables the investigation of learning-based methods for control through its computational efficiency.
Finally, addressing the sim-2-real gap is made possible by direct access to the physics engine, which allows us to improve the dynamics to match reality.

\section{Acknoledgements}
This work was partially supported by the Wallenberg AI, Autonomous Systems Software Program (WASP) funded by the Knut and Alice Wallenberg Foundation; the Stiftelsen för Strategisk Forskning (SSF) through the Swedish Maritime Robotics Centre (SMaRC)(IRC15-0046); the Swedish Defence Materiel Administration (FMV); and SAAB AB, Digital Futures and Vinnova through the ALARS project (DNR 2024-01576). 

\bibliography{references}
\end{document}